\documentclass[10pt,twocolumn,letterpaper]{article}

\usepackage{iccv}              

\definecolor{iccvblue}{rgb}{0.21,0.49,0.74}
\usepackage[pagebackref,breaklinks,colorlinks,allcolors=iccvblue]{hyperref}
 \usepackage{multirow}
 \usepackage{tabularx}
 \usepackage[table]{xcolor}
\def\paperID{6148} 
\def\confName{ICCV}
\def\confYear{2025}
\usepackage{color}
\usepackage[page]{appendix}
\usepackage{xcolor}
\usepackage{ulem}
\usepackage{utfsym}
\usepackage[accsupp]{axessibility}  
\usepackage{marvosym}

\usepackage{bbm}

\title{AdsQA: Towards Advertisement Video Understanding}

\author{Xinwei Long\textsuperscript{1}
~~~~~Kai Tian\textsuperscript{1}\thanks{Kai Tian has equal contribution with Xinwei Long.}
~~~~~Peng Xu\textsuperscript{1\Letter}
~~~~~Guoli Jia\textsuperscript{1}
~~~~~Jingxuan Li\textsuperscript{2}
~~~~~Sa Yang\textsuperscript{3}\\
~~~~~Yihua Shao\textsuperscript{4}
~~~~~Kaiyan Zhang\textsuperscript{1}
~~~~~Che Jiang\textsuperscript{1}
~~~~~Hao Xu\textsuperscript{5}
~~~~~Yang Liu\textsuperscript{2}\\
~~~~~Jiaheng Ma\textsuperscript{2}
~~~~~Bowen Zhou\textsuperscript{1,6\Letter}\\
\textsuperscript{1} Tsinghua University, \textsuperscript{2} Independent Researcher, \textsuperscript{3} Peking University, \textsuperscript{4} CASIA\\
\textsuperscript{5} Harvard University,\textsuperscript{6} Shanghai Artificial Intelligence Lab\\
    {\tt\small \{longxw22, tk23\}@mails.tsinghua.edu.cn; \{peng\_xu, zhoubowen\}@tsinghua.edu.cn}\\
{\small Project page: \url{https://github.com/TsinghuaC3I/AdsQA} 
}\\
{\small MARS2 workshop \& challenge:  \url{https://mars2workshop.github.io/iccv2025/} }
}

\begin{document}
\maketitle
\begin{abstract}

Large language models (LLMs) have taken a great step towards AGI. 
Meanwhile, an increasing number of domain-specific problems such as math and programming boost these general-purpose models to continuously evolve via learning deeper expertise.
Now is thus the time further to extend the diversity of specialized applications for knowledgeable LLMs, though collecting high quality data with unexpected and informative tasks is challenging.
In this paper, we propose to use advertisement (ad) videos as a challenging test-bed to probe the ability of LLMs in perceiving beyond the objective physical content of common visual domain.
Our motivation is to take full advantage of the clue-rich and information-dense ad videos' traits, e.g., marketing logic, persuasive strategies, and audience engagement.
Our contribution is three-fold:
\textbf{(1)}
To our knowledge, this is the first attempt to use ad videos with well-designed tasks to evaluate LLMs.
We contribute \texttt{AdsQA}, a challenging ad Video QA benchmark derived from 1,544 ad videos with 10,962 clips, totaling 22.7 hours, providing 5 challenging tasks.
\textbf{(2) }
We propose \texttt{ReAd-R}, a Deepseek-R1 styled RL model that 
reflects on questions, and generates answers via reward-driven optimization.
\textbf{(3) }
We benchmark {14}  top-tier LLMs on \texttt{AdsQA}, 
and our \texttt{ReAd-R}~achieves the state-of-the-art outperforming strong competitors equipped with long-chain reasoning capabilities 
by a clear margin.
\end{abstract}    
\section{Introduction}
\label{sec:intro}

A recent milestone flagged by OpenAI o1~\cite{jaech2024openai} and DeepSeek-R1~\cite{guo2025deepseek} has been in the spotlight, and already opened up an era of large reasoning models. 
Meanwhile, going beyond general domains, the specialized domains~\cite{zuo2025medxpertqa,ding2024automating} are increasingly studied to boost these general-purpose LLMs to evolve continuously.
The specialized domains contribute significantly in optimizing and evaluating generalist models towards specialized reasoning like domain experts~\cite{qi2024large,gao2025agent4edu}.
Recently, a clear phenomenon \cite{zhang2024accessing,guo2024deepseekcoder} has been observed that LLMs are already good at reasoning a kind of step-wise problems represented by math and programming.
As has been widely discussed~\cite{zhang2024accessing,guo2024deepseekcoder}, the step-wise problems based on explicit theorems and programming syntax are compatible with the chain-style reasoning approach \texttt{``if A, then B''}.
To further extend the diversity of specialized reasoning for LLMs, it would be better to exploit other novel domains where the domain-specific reasoning is unmanageable for {``if A, then B''}.

Therefore, we, for the first time, propose to use advertisement videos to probe the reasoning boundary of LLMs in perceiving implicit multimodal reasoning.
The domain-unique features of clue-rich and information-dense ad videos can be summarized as key words implicit, non-physical, mental, heuristic, \textit{etc}.
Different from user-uploaded videos on social media, ads are typically meticulously crafted by commercial or non-commercial organizations, making them entertaining, creative, aesthetically appealing, and capable of offering enjoyment and attracting viewer engagement.


\begin{figure*}[t]
\centering
\includegraphics[width=1\textwidth]{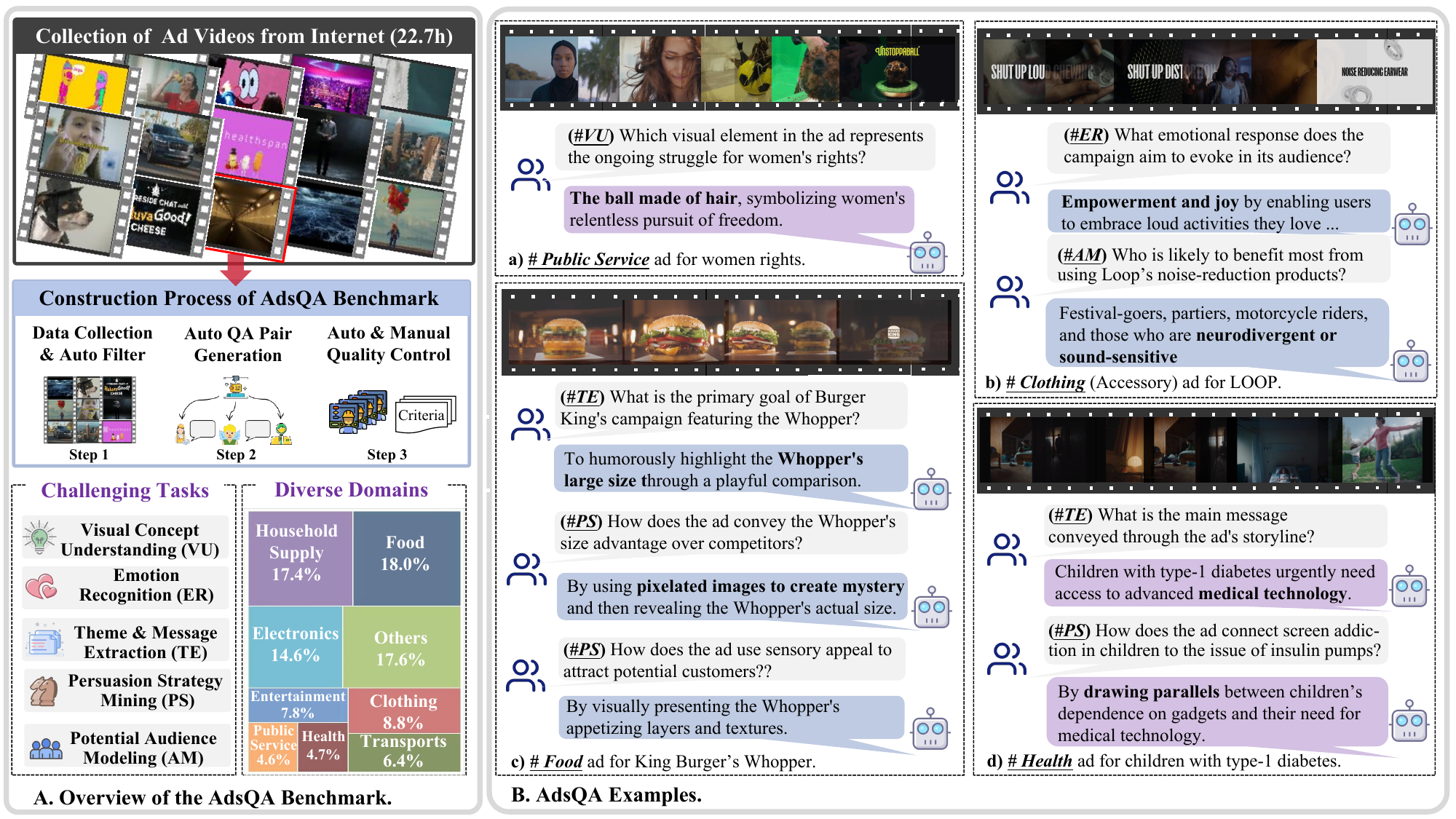} 
\caption{Overview of \texttt{AdsQA} benchmark. Subfigs A - B: statistics \& diversity, and examples. }
\label{fig2}
\end{figure*}

Based on these domain-specific advantages of ad videos,
we introduce \texttt{AdsQA}, a comprehensive and carefully curated VideoQA benchmark derived from 1,544 ad videos containing 10,962 clips, spanning a total of 22.7 hours of video content.
The \texttt{AdsQA} introduces five tasks, each requiring different types and levels of reasoning: 
(1) Visual Concept Understanding: Identify and analyze visual elements in ads.
(2) Emotion Recognition: Detect emotions and infer their roles in ads.
(3) Theme and Core Message Extraction: Summarize the central theme and key messages of the ad.
(4) Persuasion Strategy Mining: Analyze the strategies used to persuade the audience.
(5) Potential Audience Modeling: Identify and characterize the target audience.
The five tasks are formalized as the open-ended QA format. 
During the construction process, we propose an innovative Role-Played Multi-Agent Annotation framework, which simulates the role of advertising experts to generate specialized data and
significantly reduces the workload of human annotators. 
To ensure data quality, we conduct multiple rounds of automated data cleaning and manual data revision. 
Moreover, based on our \texttt{AdsQA}, we benchmark 14 well-known LLMs, including the GPT-4 level models, recently-released reasoning-based models, \textit{etc}. 

Humans process visual stimuli from ads and naturally form mental impressions, rather than relying on rigid logical reasoning templates, \textit{e.g.}, {``if A, then B''}.
Inspired by this heuristic process,
we propose \texttt{ReAd-R}, a Deepseek-R1 styled Reinforced Ad Reasoner, which learns to gather effective visual stimuli from ad videos, then reflects on the questions, and finally answers the given questions in a human-like manner.
Specifically, \texttt{ReAd-R} utilizes a reward function to evaluate the model’s responses and adjusts its parameters based on correctness, facilitating learning through trial and error.
As a result, \texttt{ReAd-R} enhances reasoning abilities through outcome-reward-driven optimization, eliminating the need for costly step-wise supervision or chain-of-thought (COT) training data.
Experiments demonstrate 
that our model brings reasoning ability an obvious gain in comprehending implicit logic of ad videos.
In contrast, other reasoning-based methods obtain limited improvement due to fixed COT templates and suboptimal process reward models.

\noindent To our knowledge, our main contribution can be stated as:\\
(1)
\texttt{AdsQA} is the first video QA benchmark for advertisement domain, which is also the first ad benchmark for LLMs, 
with domain-unique features implicit, non-physical, mental, heuristic, \textit{etc}. 
It presents a new challenge to the current mainstream \texttt{``if A, then B''} LLM thinking approach, hence further extend the domain specialization reasoning scenarios for LLMs.
Moreover, it advances ad video understanding beyond 
physical-content dominated shallow
perception towards 
deeper cognitive reasoning.\\
(2) We propose \texttt{ReAd-R}, 
a DeepSeek-R1 styled RL reasoning model, 
which enhances the specialized reasoning capability to understand 
implicit logic in ad videos
like the human thinking process from perception to cognition.
\texttt{ReAd-R} also can be regarded as one of the earliest attempts evolving R1-technique to vision research.\\
(3)
We benchmark 14  flagship LLMs on \texttt{AdsQA}, 
and our \texttt{ReAd-R}~achieves the state-of-the-art outperforming strong competitors {equipped with long-chain reasoning} (\textit{e.g.}, {variants of GPT4, LLaVA, and Qwen}) by a clear margin.
\section{Related Work}
\noindent \textbf{Video Question Answering Benchmarks.}
Video question answering (Video QA)~\cite{seo2021look,song2024moviechat+,min2024morevqa,chen2024grounded,xiong2025streaming} aims to {answer} user questions based on a given video, requiring {a detailed understanding of both the spatio-temporal} information and the relationships between objects and events~\cite{di2024grounded,xiao2024can,DBLP:journals/corr/abs-2410-10818,yao2025lens}.
{The Video QA community} has developed a variety of benchmarks. 
From the perspective of video content, most benchmarks focus on human-centric videos, such as NextQA~\cite{DBLP:conf/cvpr/XiaoSYC21}, ActivityQA~\cite{DBLP:conf/aaai/YuXYYZZT19}, MVBench~\cite{li2024mvbench}, FunQA~\cite{xie2024funqa}, and VideoMME~\cite{fu2025video}, which are derived from movies~\cite{DBLP:conf/cvpr/TapaswiZSTUF16,DBLP:journals/corr/abs-2406-10221}, TV shows~\cite{DBLP:conf/emnlp/LeiYBB18}, social media~\cite{xie2024funqa}, \textit{etc}.
Additionally, there are benchmarks centered on object-centric videos, such as EgoSchema~\cite{DBLP:conf/nips/MangalamAM23}
, which are sourced from instructional and operational videos~\cite{zhou2018towards,li2024multi}.
Despite their advancements, these benchmarks still have several limitations. MovieQA~\cite{DBLP:conf/cvpr/TapaswiZSTUF16} overly relies on dialogue understanding, which restricts the in-depth comprehension of visual content. 
Although TGIF-QA~\cite{jang2017tgif} requires some complex reasoning, the GIFs used are typically no longer than 3 seconds. 
ActivityQA~\cite{DBLP:conf/aaai/YuXYYZZT19} and Next-QA~\cite{DBLP:conf/cvpr/XiaoSYC21} have introduced open-ended QA tasks, but the annotated answers are often overly simplistic, usually consisting of just a few words. 
To sum up, current video QA benchmarks face limitations in video diversity and the setups of QA pairs. 

\noindent \textbf{Video QA Methods and Video-LLMs.} 
Early video QA methods employed graph structures~\cite{zhao2018open,jiang2020reasoning,shao2025eventvad} or transformers~\cite{li2019beyond,lei2021less,gao2023mist} to model the spatiotemporal relationships in videos and capture interactions between humans and objects. 
Recently, Video-LLMs~\cite{DBLP:conf/acl/0001RKK24,chen2023sharegpt4v,Maaz2024VideoGPT+} have demonstrated impressive capabilities in video understanding. 
Representative Video-LLMs include the VideoLLaMA~\cite{DBLP:journals/corr/abs-2406-07476,DBLP:conf/nips/MangalamAM23} series, LLaVA series~\cite{DBLP:journals/corr/abs-2311-10122,zhang2024llavanextvideo,xu2024pllava,li2024llava,liu2024llava}, InternLM series~\cite{wang2024internvideo2,internlmxcomposer2_5}, and Qwen series~\cite{DBLP:journals/corr/abs-2409-12191,bai2025qwen2}. 
Video-LLMs take both the question and the video as input and directly generate the answer, implicitly performing the reasoning and thinking process in an efficient manner.
Moreover,
Some methods attempt to explicitly model the reasoning process through step-by-step thinking or multi-round debates.
For example, some studies~\cite{fei2024video,hu2025cos,shi2024unlocking,wang2024videocot,zeng2025FSDrive} leverages chain-of-thought to simulate the thinking process, while VideoAgent~\cite{yang2024doraemongpt,fan2024videoagent,wang2024videoagent} improves reasoning capabilities through agent collaboration. 

\noindent \textbf{Advertisement Understanding.}
Internet companies generate substantial profits by automatically distributing advertisements to target users, making the understanding of ad content highly important~\cite{zahmati2023eye,jia2023kafa}.
The Pit dataset~\cite{DBLP:conf/cvpr/HussainZZYTAOK17}, one of the earliest efforts in automatic ad understanding, formalized this task as a visual question-answering (VQA) problem~\cite{long2025retrieval}. 
Subsequent research has built upon the Pit dataset or further developed it~\cite{rajakumar2024seeing}, or utilized their own private datasets~\cite{yang2024synchronized}. 
These studies have mainly focused on a single aspect of advertisement understanding, such as
persuasive strategies in image ads~\cite{ye2019interpreting,kumar2023persuasion}, image ad search~\cite{zhao2024enhancing}, intent understanding~\cite{jia2021intentonomy}, and visual metaphor comprehension~\cite{zhang2021multimet,akula2023metaclue,zhang2023multicmet,rajakumar2024seeing,xu2024exploring}.
Although the Pit dataset primarily focuses on image advertisements, it also provided a subset of video advertisements they collected. 
However, the Pit dataset suffers from issues such as data inaccessibility, lack of diversity, and limited Q\&A formats.
Therefore, there is currently no comprehensive Video QA benchmark available in the advertising domain.

\noindent \textbf{Reinforcement Learning for Reasoning.} 
Recent research efforts~\cite{jaech2024openai,jiao2024preference,luong2024reft,zuo2025ttrl} have attempted to improve the reasoning capabilities of LLMs through reinforcement learning.
Recently, DeepSeek-R1~\cite{guo2025deepseek} achieved significant improvements in reasoning based on reinforcement learning, eliminating the need for intermediate reasoning signals. 
S1~\cite{muennighoff2025s1} demonstrated that even with only a few hundred data points, the model can effectively perform reasoning in specialized tasks.
However, current research on RL-based reasoning primarily focuses on math and code problems, with few works~\cite{liu2025visual} attempting to apply it to multimodal domains, particularly open-ended video QA tasks.

\section{The \texttt{AdsQA} Benchmark}

\subsection{Task Definition}
\label{subsec3.1}
To comprehensively evaluate the model’s ability to understand ad videos, we divided the questions into five types, with each one targeting a specific angle of ad analysis. 

\noindent \textbf{Visual Concept Understanding (VU)} task evaluates the model's ability to comprehend specific visual concepts, such as characters, objects, scenes, slogans, and other details, as well as their {interrelationships}.

\noindent \textbf{Emotion Recognition (ER)} assesses if the models understand \textbf{what} emotion the ad evokes, and \textbf{how} ads establish connections with audiences through emotions.

\noindent \textbf{Theme and Core Message Extraction (TE)}  drives the models to extract the underlying message or central idea that the ad explores. 
This task requires deep reasoning to gather visual information from the full video.

\noindent \textbf{Persuasion Strategy Mining (PS)} task evaluates the model's ability to uncover the strategies used to convey core messages and persuade audiences, such as humor, exaggeration, and visual rhetoric. The task includes questions, such as: \textbf{how} an ad conveys its central message, \textbf{why} the ad is appealing, and \textbf{what} strategies the ad employs. 

\noindent \textbf{Potential Audience Modeling (AM)} probes the model's performance in identifying potential audience groups and their profiles. It sets key questions, such as \textbf{who} the ad targets and \textbf{what} the characteristics of the audience are. This task directly reflects the ad's influence and value.

These five tasks are defined as the open-ended video QA and each question is annotated with a ground-truth answer. 

\subsection{Dataset Construction Pipeline}
\texttt{AdsQA} benchmark construction pipeline is in three stages: 

\label{subsec3.2}
\noindent \textbf{Pre-Processing.} To serve the ad video understanding, we consider the creativity, aesthetic quality, and availability of videos during our collection process.
We do not use any private data; instead, we collect creative ad videos that adhere to the Creative Commons License from a creative community platform~\cite{adsw}. 
This platform offers publicly accessible, high-quality ad videos uploaded by creators.
Specifically, we first crawl videos along with their metadata.
The metadata is typically written by the uploader (\textit{i.e.}, usually the creator of the ad) and includes information such as the theme, content, and key creative elements of the ad. 
It can be considered the ``ground-truth'' information of the ad video and serves as an important reference for our benchmark construction.
Then, human experts are asked to carefully review the completeness and accuracy of the metadata and manually remove incomplete samples. 
Additionally, we conduct automated filtering based on video duration, aesthetic score, and content to ensure video quality and exclude non-ethical and negative content.

Considering efficiency and accuracy, we employ PySceneDetect to segment ad videos into fine-grained video clips to avoid missing detailed information in subsequent steps. 
For each clip $C_i$, we use Video-LLMs to generate its description $Desc_i$. 
To preserve key visual elements, we sample $n$ keyframes $\{F_i^0,..., F_i^{N_{max}}\}$ from each clip based on frame similarity calculated using SSIM~\cite{wang2004image}, where $n$ is determined by the duration of each clip. 
To retain speech information, we extract speech information $Asr_i$ for each clip using the Whisper model, and translate them into English using GPT-4~\cite{hurst2024gpt}.
Therefore, an ad video with $N$ clips can be represented as a modality-interleaved sequence, as Eq.~\ref{eq1},
\begin{equation}
V = \{M, \{F_i^0,..., F_i^{n_{i}}, Desc_i, Asr_i\}_{i=1}^{N} \},
\label{eq1}
\end{equation}
where $M$ denotes the meta-information of the video. The modality-interleaved sequence will be used for automated annotation generation.

\noindent \textbf{Role-Played Multi-Agent Annotation.}
To balance data quality and cost, ensure data diversity, and avoid template-driven outputs, we propose a Role-Played Multi-Agent Annotation framework to automatically generate video QA pairs.
Inspired by previous research~\cite{yuan2024evoagent}, we found that agents can develop specialized capabilities in the ad domain by configuring specific skill sets, such as marketing, visual design, and consumer psychology. 
Therefore, we can leverage AI agents to act as ad experts, analyzing diverse facets of ad videos and designing more specialized and challenging questions. 
We ask human experts to create profiles for advertising expert skills, enabling AI agents to adaptively select profiles (or autonomously generate new profiles) to accomplish role-playing tasks.
Additionally, human experts provide examples of QA pairs as in-context demonstration to guide the agents in generating appropriate QA pairs.

The Role-Played Multi-Agent Annotation is formulated into a three-stage pipeline.
(1) In the initial stage, a master agent is recruited to generate a preliminary QA annotation for the given modality-interleaved sequence $V$.
(2) In the iterative stage, the master agent checks the quality of the current QA annotations and decides whether to recruit a specialized expert agent.
If necessary, the master agent selects a profile for the expert agent and instantiates it accordingly.
The expert agent then generates new QA annotations leveraging its specialized expertise.
Finally, the master agent revises the original annotations based on the expert agent's provided annotations and assesses whether to terminate the iteration or continue to recruit a new expert agent.
(3) After terminating the iteration, the master agent synthesizes outputs from previous iterations to produce the final QA annotation.

\noindent \textbf{Automated Data Cleaning.}
After automated annotation through multi-agent collaboration, 
we employ the IXC-2.5-Reward model~\cite{zang2025internlm} to automatically verify the correctness of the answers. Given the video, question, and meta information, the reward model scores the answer. QA pairs with scores below a certain threshold are flagged and carefully reviewed by annotators.

\subsection{Manual Check and Quality Control}
\label{subsec3.3}
Following the automatic annotation generation, we conduct a careful manual check and revision on all Q\&A pairs, focusing on the question suitability, annotation correctness, video content, and task difficulty.
To ensure the quality of Q\&A pairs, we first remove questions that are homogeneous, unrelated to the ads' theme, or cannot be answered based on the ad video (\textit{e.g.}, those requiring external knowledge).
Then, we incorporate the meta information as ground truth to verify the accuracy of the reference answers. We eliminated pairs containing inaccurate, ambiguous, or biased answers. 
After two rounds of rigorous selection, only 37\% of the QA pairs remain.
Additionally, we manually revise the wording of some QA pairs to improve their clarity and accuracy. 
Any modifications to a sample are reviewed by other annotators to ensure consistency and reliability.

\subsection{Dataset Statistics}
\label{subsec_bench}

\begin{figure*}[t]
\centering
\includegraphics[width=0.9\textwidth]{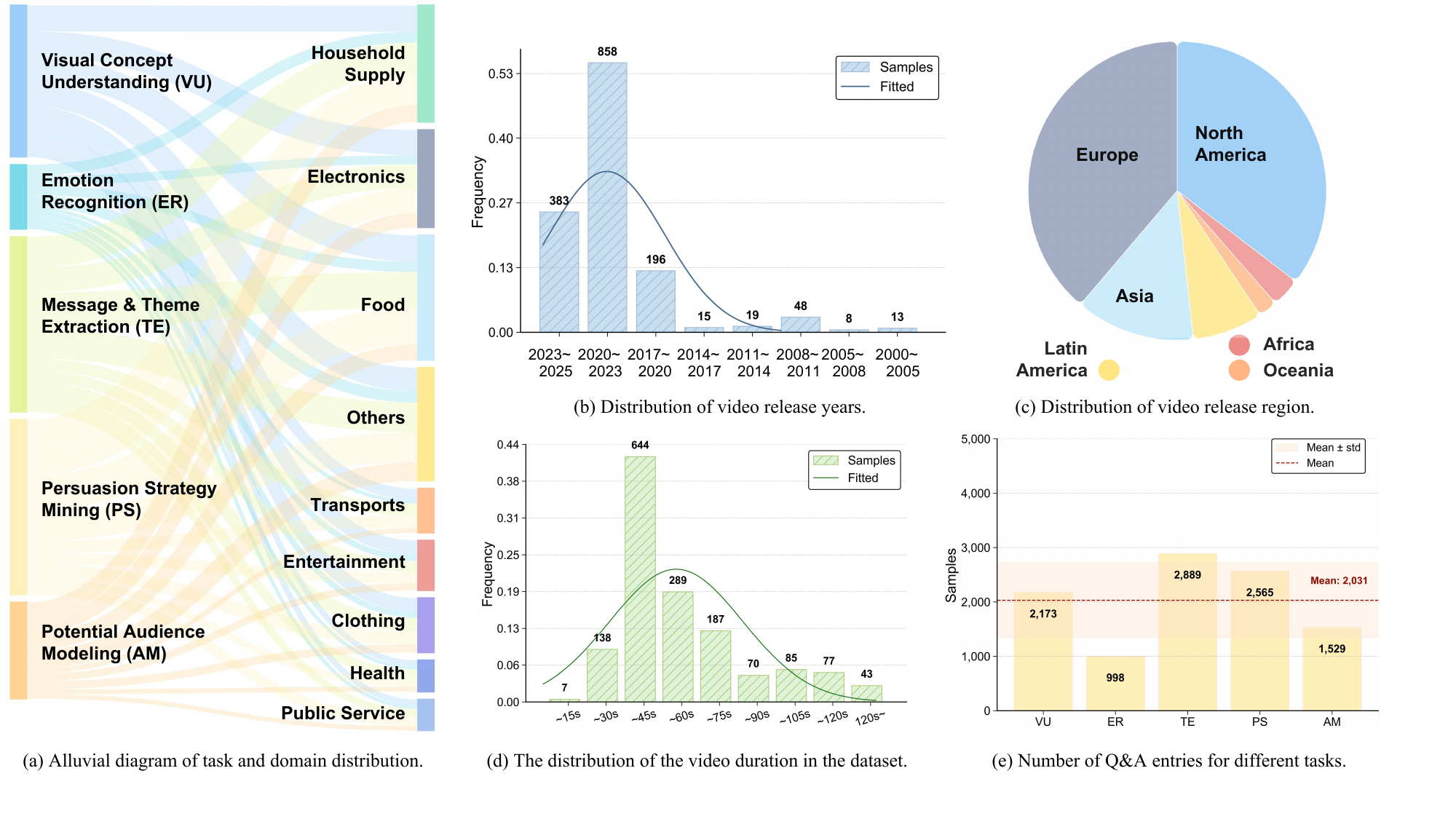} 
\caption{Statistics of Our \texttt{AdsQA} Bench. 
}
\label{fig:bench}
\end{figure*}

\noindent \textbf{Videos.}
Our \texttt{AdsQA} Benchmark includes a total of 1,544 unique advertisements, comprising 10,962 video clips for evaluation. 
As shown in Fig.~\ref{fig:bench}a, these ad videos span 9 primary domains.
The video durations range from 15 to 120 seconds, with an average length of 52.9 seconds, amounting to a total of 22.7 hours of content. 
Each ad video is accompanied by metadata, including a title, tags, automatic speech recognition (ASR) results, and descriptions. This metadata is utilized for Q\&A generation, as described in Sec. ~\ref{subsec3.1}. 
Most of the ad videos are in English and primarily originate from North America and Europe. 

\noindent \textbf{Q\&A Pairs.} 
Our \texttt{AdsQA} Benchmark includes 7,859 QA pairs (among which 29.2\% of the questions can be categorized into two types, resulting in a total of 10,154 QA pairs for evaluation).
Among these:
Visual Concept Understanding accounts for 21.4\% of the QA pairs,
Theme and Message Extraction accounts for 28.5\%,
Persuasion Strategy Mining accounts for 25.3\%,
Potential Audience Modeling accounts for 15.1\%, and
Emotion Recognition accounts for 9.8\%.
Typical persuasion strategies include metaphor, symbolism, humor, expert opinion, and others. 
Some questions may fit into two categories, such as  ``What is the role of a particular visual concept in relation to the theme?". 
On average, each question consists of 17 words and each answer contains approximately 12 words.


\section{Methodology}
\label{sec:methodology}
\begin{figure}[t]
\centering
\includegraphics[width=0.45\textwidth]{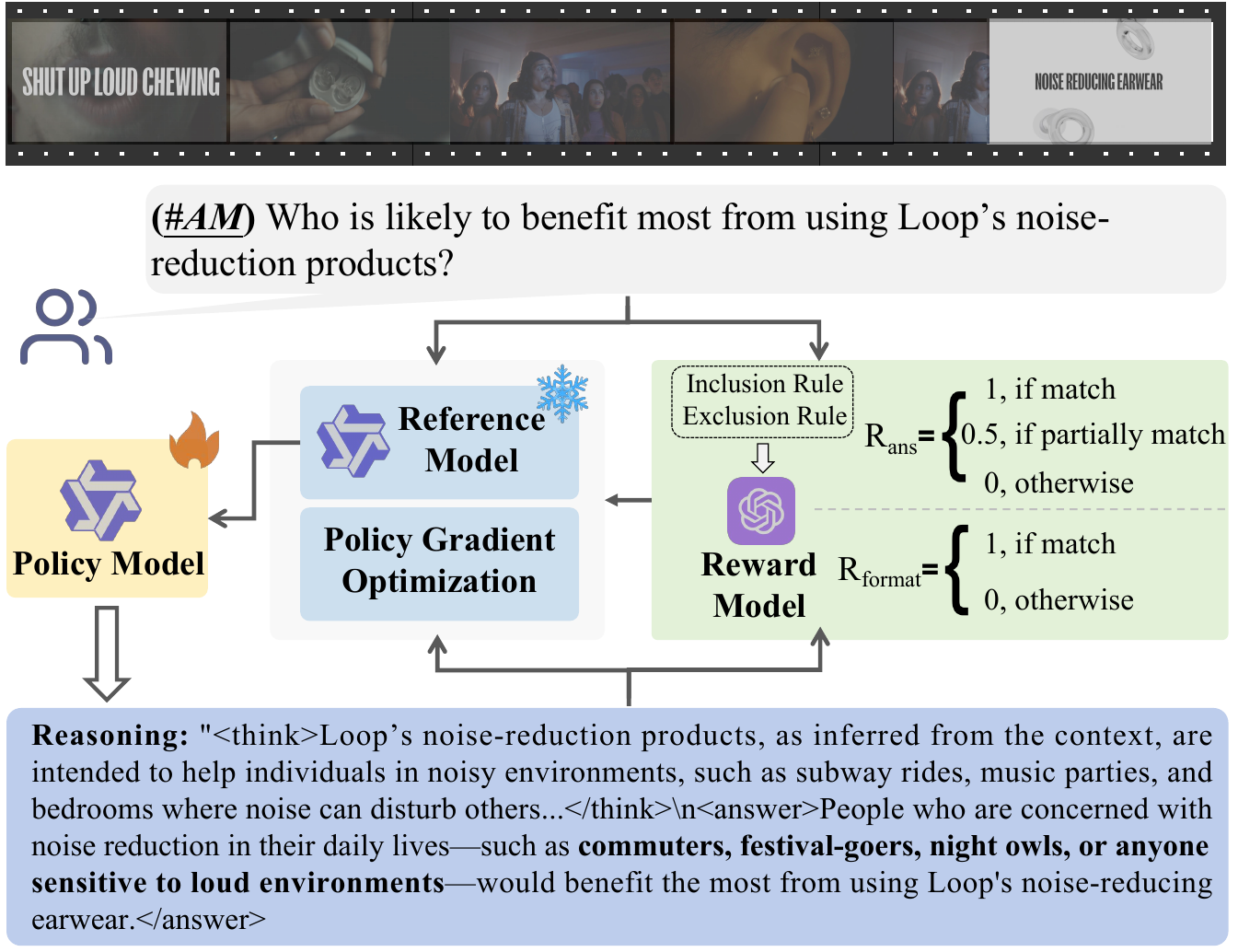} 
\caption{Framework of \texttt{ReAd-R}. Given a question and video, the policy model generates multiple responses. The reward model evaluates and scores them, and the rewards are used to update the policy model via policy gradient optimization.}
\label{fig:read}
\end{figure}
As illustrated in Fig.~\ref{fig:read}, we propose \texttt{ReAd-R}, a DeepSeek-R1 styled {Reinforced Ad Reasoner}, to simulate human heuristic thinking and learn from trial and error.
The model takes an ad video and a question as input. 
The policy model generates a set of reasoning processes, including thoughts and answers, based on the given input.
Each reasoning process is evaluated using a reward function to compute its reward value.
After calculating the reward values for all outputs, each reasoning process is assessed and used to update the policy model.
Additionally, \texttt{ReAd-R} employs KL divergence to control the difference between the policy model and the reference model, ensuring stable training.

\noindent \textbf{Data Preparation.} 
Inspired by previous work~\cite{muennighoff2025s1,guo2025deepseek}, RL-based reasoners can improve reasoning capabilities with limited high-quality data. 
We additionally crawled a total of 1,053 ad videos along with their metadata and automatically generated annotations using the same method. 
We utilized metadata tags collected during data crawling (\textit{e.g.}, topic, domain) to ensure diversity, adopted Video-LLM evaluation results to assess difficulty, and performed manual quality assessment based on the same criteria as in Sec.~\ref{subsec3.3}.
Finally, we selected 100 videos and 500 question-answer pairs.
We designed a prompt format to guide the model to output its reasoning process before providing the final answer.

\noindent \textbf{Reward Modeling.} 
\label{reward}
The reward model guides the model to find the optimal direction through trial and error.
Ideally, \texttt{ReAd-R}'s responses should (1) include as many elements of the standard answer as possible and (2) avoid containing content not mentioned in the ad video (\textit{e.g.}, hallucinations).
To achieve this, we propose a rule-guided LLM evaluator as the reward model to assess the quality of each response.
We introduce two sets of rules: inclusion rules and exclusion rules.
The inclusion rule requires that the generated answer incorporate as many elements of the standard answer as possible. If satisfied, the reward value is 1.0; otherwise, it is 0.
The exclusion rule states that if the generated content includes elements not mentioned in the standard answer and these elements cannot be inferred from the meta-information, they should be judged as incorrect.
To avoid sparse reward values in the early stages, we relaxed the inclusion rule. If the generated response partially includes factual elements, it is assigned a reward value of 0.5.
Additionally, the format reward is used to enforce the model's predictions to adhere to the required format of $<$think$>$ and $<$answer$>$, as $R(\cdot)=R_{ans}(\cdot) + R_{format}(\cdot)$.

\noindent \textbf{Reinforced Fine-tuning.} 
Following DeepSeek-R1, we employ the GRPO algorithm instead of PPO to optimize our model.
\texttt{ReAd-R} first generates $n$ distinct responses $O=\{o_1,o_2,...,o_n\}$ from the current policy model $\pi_{\theta
_{old}}$.
Then, the reward model $R(\cdot)$ evaluates these responses $O$ to obtain their reward value as $\{r_1,r_2,...,r_n\}$,
and the advantage $A_i$ is defined as
\begin{equation}
A_i =\frac{r_i - {\rm mean}(\{r_1,r_2,...,r_n\})}{{\rm std}(\{r_1,r_2,...,r_n\})}.
\label{eq2}
\end{equation}
$A_i$ denotes relative quality of the $i^{th}$ answer when compared to the group of rewards $\{r_1,r_2,...,r_n\}$.
This strategy encourages the model to generate better answers with higher reward values within the group.
We use GRPO algorithm~\cite{guo2024deepseekcoder} to optimize our policy model based on $A_i$. 


\section{Experiments}

\subsection{Experimental Settings}
\noindent \textbf{Evaluated Models.}
We select 14 top-tier LLMs as strong competitors covering diverse categories:
(1) GPT-4 Level Baselines: including GPT-4o~\cite{hurst2024gpt}, GPT-4V~\cite{achiam2023gpt} and Qwen2.5-VL-72B~\cite{bai2025qwen2}.
(2) Open-sourced Video LLMs: We present at least one representative model released in the past six months from each series of Video-LLMs, such as Qwen2.5-VL-7B~\cite{bai2025qwen2}, LLaVA-Onevision-7B~\cite{li2024llava}, and others. 
(3) Reasoning Based Methods: We re-implemented several reasoning-based methods and applied them to the \texttt{AdsQA} task. These methods include the Video Chain-of-Thought (VOT)~\cite{fei2024video}, the Role-Played Agent (EvoAgent)~\cite{yuan2024evoagent}, and the Monte Carlo Tree Search (MCTS) algorithms~\cite{zhang2024accessing}.
Detailed descriptions and implementation methods are provided in the appendix.

\noindent \textbf{Evaluation Metric.}
All five tasks of \texttt{AdsQA} are evaluated as open-ended QA, allowing for the free-text evaluation methods.
Traditional free-text metrics~\cite{lin2004rouge,papineni2002bleu} are sensitive to lexical variations, which may introduce bias when evaluating different LLMs.
Recent studies~\cite{chiang2023can,xie2024funqa,DBLP:journals/corr/abs-2406-10221,chen2023autoeval} have shown promising results in using large generalist models to evaluate generated text. 
Therefore, we follow their framework and employ GPT-4o to assist in evaluating free-text similarity.
Specifically, we use the same inclusion and exclusion rules mentioned in Sec.~\ref{reward} to guide the scoring of text generated by the large model.
We provided each sample with a ground-truth answer and meta-information; each meta-information includes relevant information such as the creative elements, content, storyline, and themes of the advertisement video.
We propose both relaxed and strict scoring modes. 
Our \textbf{strict accuracy} \( Acc_{\text{strict}} \) is defined as:
\begin{equation}
        \text{Acc}_{\text{strict}} = \frac{1}{n} \sum_{i=1}^n \mathbbm{1}\{x_i = y_i\},
        \label{eq:strict}
    \end{equation}
where {\( \mathbbm{1}\{\cdot\} \)} is the indicator function that returns 1 if the model's response \( x_i \) semantically matches all elements of the standard answer \( y_i \), and 0 otherwise.
    
In the relaxed mode, the \textbf{relaxed accuracy} \( Acc_{\text{relaxed}} \) is defined as:
\begin{equation}
\begin{split}
    \text{Acc}_{\text{relaxed}} &= \frac{1}{n} \sum_{i=1}^n (\mathbbm{1}\{x_i = y_i\} + \lambda \cdot \mathbbm{1}\{x_i \approx y_i\}),
    \label{eq:relaxed}
\end{split}
\end{equation}
where \( \mathbbm{1}\{x_i \approx y_i\} \) returns 1 if the response \( x_i \) partially addresses the elements of \( y_i \), and 0 otherwise. 
It is observed $\lambda=0.5$ works well in experiments.
Our prompt template for model-based evaluation are in the appendix.

\begin{table*}[t]
\small
\centering
\resizebox{\linewidth}{!}{
\begin{tabular}{@{}ccccccccccccc@{}}
\toprule
\multirow{2}{*}{\textbf{Model}} & \multicolumn{6}{c}{\textbf{Strict Accuracy}} & \multicolumn{6}{c}{\textbf{Relaxed Accuracy}} \\ \cmidrule(r){2-7} \cmidrule(l){8-13}
                       & \textbf{VU}  & \textbf{ER}  & \textbf{TE}  & \textbf{PS} & \textbf{AM} & \textbf{Overall} & \textbf{VU}  & \textbf{ER}  & \textbf{TE}  & \textbf{PS}  & \textbf{AM} & \textbf{Overall} \\ \midrule
Human Results$*$ & 55.1 & 39.4 & 57.1 & 57.5 & 45.8 & 51.3 & 74.5 & 65.9 & 75.7 & 76.2 & 64.6 & 71.4   \\                    \midrule
\rowcolor{gray!10} \multicolumn{13}{c}{Commercial Large
Muitimodal
Model}                                                                           \\ \midrule
 GPT-4o~\cite{hurst2024gpt}                     & 24.9   & 26.5   & 32.6   & 32.3  & 31.0  & 29.4       & 50.1   & 57.8   & 62.4   & 55.2  & 54.8  & 56.6        \\
 Gemini-2.5-Pro$*$~\cite{comanici2025gemini}   & 36.0 & 44.0 & 40.4 & 29.3 & 33.3  & 35.9       & 59.2  & 69.0  & 64.0  & 54.4  & 62.4   & 60.7        \\
 \midrule
\rowcolor{gray!10} \multicolumn{13}{c}{Open-sourced
Video-LLMs}                                                                           \\ \midrule
VideoLLaMA2-7B~\cite{DBLP:journals/corr/abs-2406-07476}                    & 4.56   & 7.75   & 7.28   & 6.06  & 7.38  & 6.48       & 21.6   & 29.0   & 28.4   & 22.6   & 27.2  & 25.2       \\
InternLM-XComp.2.5-7B~\cite{internlmxcomposer2_5}                     & 11.2   & 10.7   & 16.3   & 11.9  & 12.1  & 12.6       & 34.7   & 35.9   & 41.5   & 34.1   & 37.3  & 36.5       \\
LLaVA-OneVision-7B~\cite{li2024llava}                     & 11.8   & 11.6   & 16.2   & 13.7  & 15.1  & 14.0       & 35.8   & 39.5   & 43.2   & 36.8   & 41.7  & 39.1       \\
LLaVA-Video-7B~\cite{zhang2024video}                     & 14.0   & 14.4   & 18.7   & 15.2  & 17.3  & 16.1       & 37.8   & 41.6   & 45.1   & 38.1   & 43.6  & 41.0       \\
MiniCPM-o\_2.6-7B~\cite{yao2024minicpm}                     & 13.0 & 15.3 & 17.3 & 14.9 & 18.9  & 16.3       & 38.1  & 43.4  & 45.7  & 39.1  & 45.0   & 42.5      \\
\textbf{Qwen2-VL-7B}~\cite{DBLP:journals/corr/abs-2409-12191}                     & \textbf{13.7} & \textbf{15.4} & \textbf{20.2} & \textbf{16.0} & \textbf{20.0} & \textbf{17.2} & \textbf{36.8} & \textbf{41.3} & \textbf{46.0} & \textbf{37.7} & \textbf{44.6} & \textbf{41.0}       \\
Qwen2.5-VL-7B~\cite{bai2025qwen2}                     & 20.2 & 23.2 & 24.6 & 20.5 & 24.3  & 23.0       & 45.8  & 50.4  & 51.8  & 45.3  & 50.9  & 48.9       \\
 Qwen2.5-VL-72B~\cite{bai2025qwen2}                     & 26.7   & 30.2   & 34.8   & 29.8  & 34.7  & 31.0       & 51.8   & 57.4   & 59.5   & 53.8   & 59.2  & 55.8       \\
 \midrule
\rowcolor{gray!10} \multicolumn{13}{c}{Reasoning-based Models}                                                                           \\ \midrule
VOT (Qwen2-VL-7B)~\cite{fei2024video}                     & 14.3   & 16.4   & 19.5   & 12.6  & 22.0  & 17.0      & 36.5   & 43.3   & 42.5   & 35.2   & 45.1  & 40.2       \\
EvolAgent (Qwen2-VL-7B)~\cite{yuan2024evoagent}                     & 15.0   & 16.1   & 21.6   & 16.5  & 20.4  & 18.3       & 38.6   & 43.4   & 46.4   & 40.4   & 45.3  & 42.6       \\
MCTSr (Qwen2-VL-7B)~\cite{zhang2024accessing}                  & 14.8   & 10.2   & 19.3   & 16.1  & 16.4  & 17.1       & 40.6   & 41.3   & 46.5   & 41.1   & 42.6  & 43.4       \\ \midrule
Qwen2-VL-7B (SFT)~\cite{DBLP:journals/corr/abs-2409-12191}                     & 10.5   & 12.5   & 17.9   & 13.8  & 15.9  & 14.1       & 34.8   & 42.1   & 42.7   & 34.9   & 41.4  & 38.8       \\
 \textbf{ReAd-R (Qwen2-VL-7B) (Ours)}                  &  \textbf{15.8} & \textbf{19.3} & \textbf{21.7} & \textbf{18.5} & \textbf{18.8} & \textbf{18.6} & \textbf{42.3} & \textbf{46.4} & \textbf{50.0} & \textbf{43.4} & \textbf{47.8} & \textbf{44.9}       \\
ReAd-R (Qwen2.5-VL-7B) (Ours)                & 20.4 & 27.9 & 27.2 & 22.1 & 25.5 & 25.0 & 46.2 & 56.2 & 54.6 & 48.2 & 52.6 & 51.5       \\ 
 \bottomrule
\end{tabular}
}
\caption{Experimental Results.}
\end{table*}

\noindent \textbf{Implementation Details.} 
Our \texttt{ReAd-R} is model-agnostic. In our main experiments, we utilize Qwen2-VL-7B as the base model.
We froze the parameters of the visual backbone and only fine-tuned the parameters of the language model. We conducted the experiments with a batch size of 4 and a learning rate of 1e-6 on a server with eight A100 GPUs, with a training duration of 12 hours. Additional details and hyperparameters are provided in the appendix.



\subsection{Results and Observations}

Although GPT-4 achieves over 85\% accuracy on other datasets such as Next-QA, its strict accuracy on our dataset is only 29.4\%, and its relaxed accuracy is just 56.6\%. 
This shows that our benchmark exceeds the capabilities of some multi-modal large models.
We observe that the state-of-the-art multi-modal models show strong visual perception, often generating partial answers by describing video segments.
However, fully understanding the underlying meaning of ads is difficult for all baseline models. This requires not only identifying specific visual elements but also interpreting their implicit logic.

Different tasks vary in difficulty, but their performance is comparable.
Though tasks focus on different aspects, these aspects are closely linked. For example, ads may use emotions or specific visuals to convey themes.
Most models perform best on Theme and Message Extraction and Audience Modeling tasks. 
This is because ad videos aim to deliver clear messages to specific audiences, making it relatively easier to identify themes and target users.
Even without fully understanding creative elements or marketing strategies, models can infer these from characters, scenes, or slogans in the ad.
The Persuasion Strategy Mining task is more challenging. It requires models to explain how and why ads use certain designs or elements.
These questions are often not directly expressed in the video and may even require interpreting counterfactual or unexpected visual information, which increases the difficulty of this task.
Models also perform worse on the Visual Concept Understanding (VU) task. While other video QA benchmarks~\cite{DBLP:conf/aaai/YuXYYZZT19} focus on surface-level questions like ``What color are the gloves?'', our VU task requires understanding global ad information, such as ``Which scene represents the ad's theme?''
Answering this question requires understanding the theme, identifying related scenes, and describing them briefly.

On the \texttt{AdsQA} task, reasoning models that excel in code and math tasks achieved only marginal results, even with ten times the computational cost. This shows that \texttt{AdsQA} reasoning differs significantly from tasks like math and code.
Math and code rely on structured ``if A, then B'' reasoning, while ads require associative reasoning, connecting concrete visuals to abstract concepts.
Prior work~\cite{zhong2024let} also suggests chain-of-thought reasoning can hinder abstract reasoning performance.
Though MCSTr improved results by 2.4\% over Qwen2-VL, its search process is inefficient and highly dependent on the reward model. 
In some cases, Qwen2-VL answers correctly, but MCSTr fails due to incorrect search directions from the reward model.
Moreover, training an effective reward model is difficult, especially in the data-scarce ad domain.

Our method, \texttt{ReAd-R}, achieves a 3.9\% improvement using only 500 video-question-answer triplets, outperforming Qwen2-VL and other reasoning models.
During inference, our model reasons about video content and questions by generating free-text explanations, eliminating the need for fixed chain-of-thought (COT) templates (\textit{e.g.}, VOT), additional reward functions (\textit{e.g.}, MCTSr), or iterative search processes (\textit{e.g.}, MCTSr and EvolAgent).
Interestingly, we observe that supervised fine-tuning (SFT) on the same data caused a sharp performance drop of 2.2\% for Qwen2-VL. 
Using limited but high-quality data does not improve the performance of SFT models. SFT may overfit to the features of the training data, leading to poor generalization on diverse test sets.
In contrast, our \texttt{ReAd-R} model improves reasoning capabilities through RL on limited data, achieving generalization across diverse advertisement videos.

\subsection{Ablation {Study}}
\begin{table}[]
\centering
\footnotesize
\resizebox{0.9\linewidth}{!}{
\begin{tabular}{ccc}
\toprule
                        & Strict Acc. & Relaxed Acc. \\ \midrule
\texttt{ReAd-R} (Our Model)     & \textbf{18.6}            & \textbf{44.9}             \\ \midrule
w. Uncurated Data       & 13.2             & 36.2              \\
w. Strict Reward        & 15.1             & 41.7              \\
w.o. Prompt Constraints & 16.8           & 42.9             \\ \midrule
Qwen2-VL (SFT) & 14.1             & 38.8              \\ 
w. Uncurated Data & 16.0            & 40.9              \\ \midrule
VOT (Chain length: 3) & 17.0             & 40.2              \\ 
w. (Chain length: 5) & 13.6             & 36.2              \\
\bottomrule
\end{tabular}
}
\caption{Ablation study on \texttt{ReAd-R}, {Qwen2-VL SFT}, and VOT.}
\label{ablation}
\end{table}
We conduct ablation studies for our \texttt{ReAd-R} model in Tab~\ref{ablation}. 
We first show results of RL fine-tuning on 1,053 uncurated videos and 5,159 QA pairs.
Our model performs worse than Qwen2-VL.
This is due to the GRPO algorithm's sensitivity to data quality. Early in fine-tuning, low-quality data can mislead optimization, causing suboptimal results.
Next, we test a strict reward rule (only fully correct answers receive a {$R_{ans}$} reward of 1; otherwise, the reward is 0). 
Our method does not achieve improvement through RL fine-tuning.
During training, we observe that the strict reward rule makes it hard for the model to earn rewards, limiting gains.
We also find \texttt{ReAd-R} is sensitive to prompt templates. Properly constraining the model's thinking and answers (\textit{e.g.}, avoiding repetitive content in $<$think$>$ $<$/think$>$ and $<$answer$>$ $<$/answer$>$ tags) stabilizes training and improves performance.

We also conduct ablation studies on our implemented baselines (\textit{e.g.}, VOT, {Qwen2-VL SFT}). We present the results of Qwen2-VL fine-tuned using uncurated data.
Interestingly, the model fine-tuned on uncurated data improves by 2.1\% over limited-data fine-tuning but still lags behind the non-SFT model. This shows SFT does not enhance reasoning on ad data.
Additionally, we experiment with VOT using a COT length of 5 (as set in its original paper) and observe a significant performance drop of 4.0\%. The model does not benefit from step-by-step reasoning; instead, accumulated errors degrade its performance. This highlights the difference between abstract reasoning in the ad domain and chain-style logical reasoning.

\begin{figure}[t]
\centering
\includegraphics[width=0.45\textwidth]{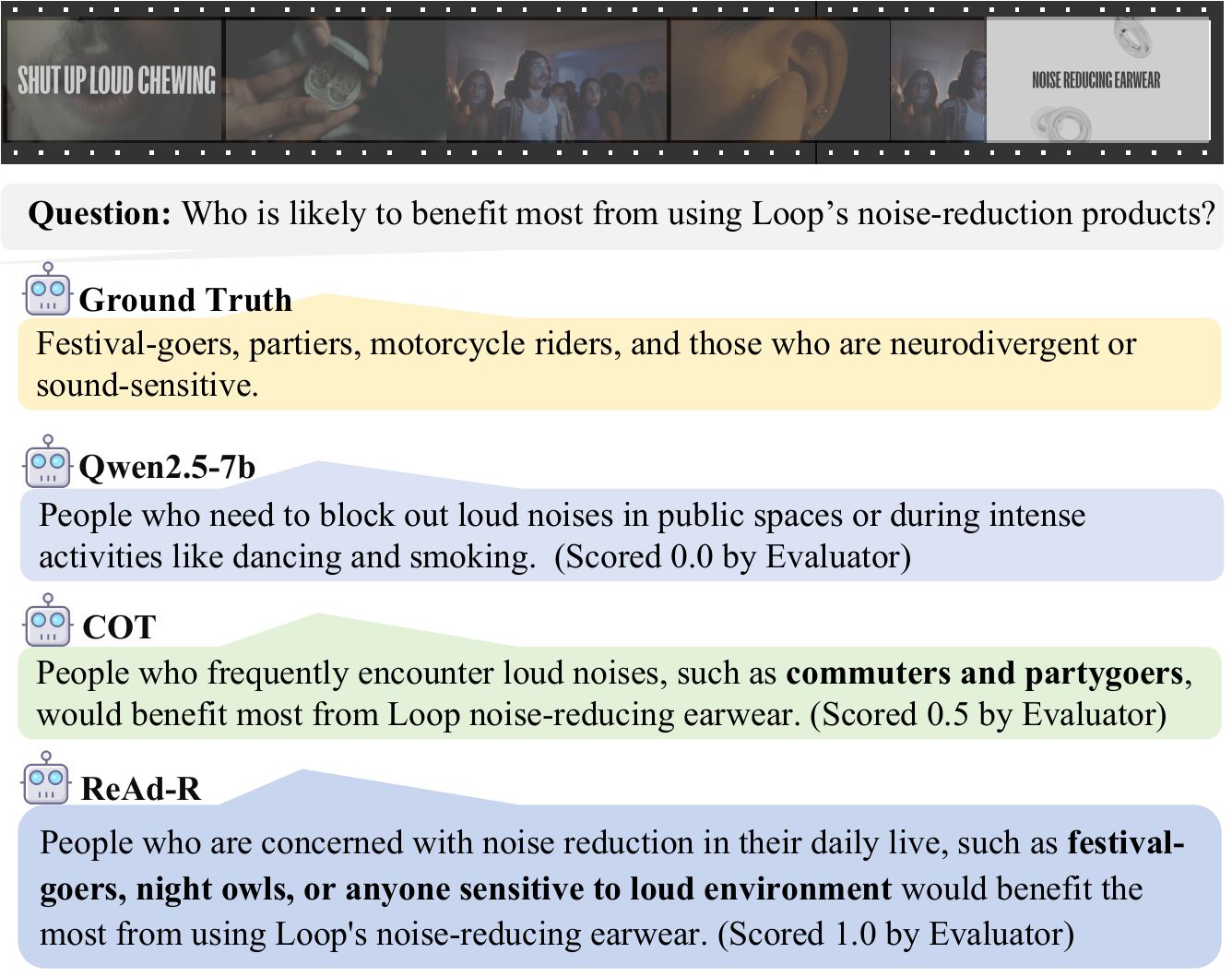} 
\caption{Case Study. Due to the limited page space, we omit the model's thinking process and show only final answer.}
\label{fig:cs}
\end{figure}

\subsection{Case Study}
As shown in Fig.~\ref{fig:cs},
Qwen2.5-VL generates vague and overly general answers (scored 0 by GPT-4). The COT baseline produces better answers (scored 0.5 by GPT-4) by gathering information step-by-step through a reasoning process, but it still misses some important clues in the video. 
Our \texttt{ReAd-R} learns to perform video-grounded reasoning through reinforcement fine-tuning, deriving the final answer by summarizing its reasoning process. 
The reinforced thinking-answer mechanism helps \texttt{ReAd-R} generate more complete and precise answers.


\section{Conclusion}

We contibute \texttt{AdsQA}, the first advertisement benchmark for LLMs, 
with domain-unique features implicit, {non-physical}, mental, heuristic, \textit{etc}. 
It presents a new challenge to the current  ``if A, then B'' LLM thinking approach, hence further extend the domain specialization reasoning scenarios for LLMs.
We propose  \texttt{ReAd-R}, 
a DeepSeek-R1 styled RL reasoning model, no need of costly supervision, 
which enhances the specialized reasoning capability to understand 
implicit logic in ad videos.
We benchmark 14  flagship LLMs on {\texttt{AdsQA}}, 
and our {\texttt{ReAd-R}}~achieves the state-of-the-art outperforming strong competitors equipped with long-chain reasoning capabilities (\textit{e.g.}, {variants of GPT4, LLaVA, and Qwen}).

\noindent {\bf Acknowledgments.}
This work was supported by the National Key Research and Development Program of China (2022ZD0160603), and NSFC No.62306162.

\normalem
{
    \small
    \bibliographystyle{ieeenat_fullname}
    \bibliography{main}
}
\clearpage
\appendix

\section{Benchmark Construction}
\subsection{Pre-processing}
Our pipeline begins with collecting Creative Commons licensed videos from a professional advertising platform, focusing on high-quality, award-winning content, including Clio Award recipients. 
Each video is crawled with its original metadata that contains {creator-specified} information about themes, content, and key creative elements, {which are used as our ground truth for benchmark construction}. 
{Our human annotators manually review the metadata to verify completeness and accuracy, and remove the incomplete samples.} 
We then perform automated filtering based on three criteria: video {length (retaining 15s-150s clips)}, aesthetic quality (using computational scoring), and content appropriateness (removing non-ethical material).
For temporal analysis, we employ PySceneDetect to segment videos into coherent clips while preserving narrative flow. 
From each clip $C_i$, we extract key visual elements by sampling $n$ keyframes $\{F_i^0,...,F_i^n\}$ based on SSIM-calculated frame similarity, where $n$ adapts to clip duration. 
Complementary multimodal features are obtained through: (1) Video-LLM generated descriptions ($Desc_i$) capturing visual content, and (2) Whisper-based speech transcription ($Asr_i$) with GPT-4 translation for English standardization. 

\begin{table*}[htbp]
\centering

\begin{tabular}{cc}
\toprule
\textbf{Domains} & \textbf{Sub-domains} \\
\midrule
Foods &  \#Wine\&Spirits \#Soft Drinks \#Alcoholic Drinks \#Snacks   \\
Electronics &  \#Media \#Electronics Technology \#Digital Gaming \#TV\&Streaming \#Music  \\
Health & \#Health Care \#Pharmaceutics \\
Household Supply & \#Retail Service \#House\&Garden \#Pets \#Education \#Household products\\
Public Service &  \#Protection of Rights \#Public Service Announcements\\
Entertainment &   \#Travel\&Tourism \#Festival\&Event \#Holiday \#Recreation Leisure \#Hospitality\\
Transport & \#Public Transport \#Automotive \\
Clothing, Fashion, Sports \& Accessory & \#Beauty \#Accessory\&Jewelry \#Sportswear \#Fashion\\
Others &  \#Public Interest \#Industrial \#Professional Service \#Office Equipment \\
\bottomrule
\end{tabular}
\caption{Theme taxonomy of \texttt{AdsQA} videos, covering nine domains and 33 sub-domains. \# indicates individual sub-domain hashtag.}
\label{tab:domains}
\end{table*}

\subsection{Role-Played Multi-Agent Annotation}
\begin{figure}[t]
\centering
\includegraphics[width=0.5\textwidth]{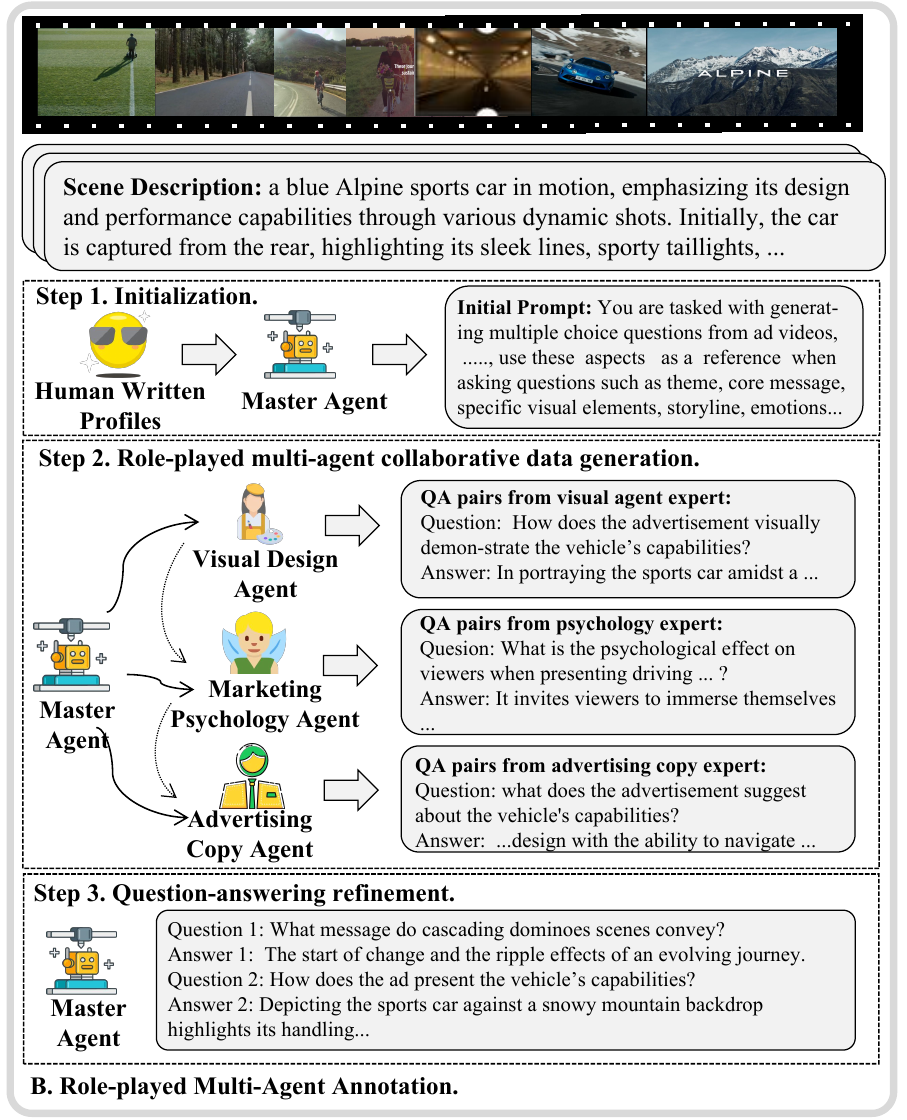} 
\caption{Role-played multi-agent processing pipeline.}
\label{fig_1}
\end{figure}

\begin{figure*}[htbp]
\centering
\includegraphics[width=1.0\textwidth]{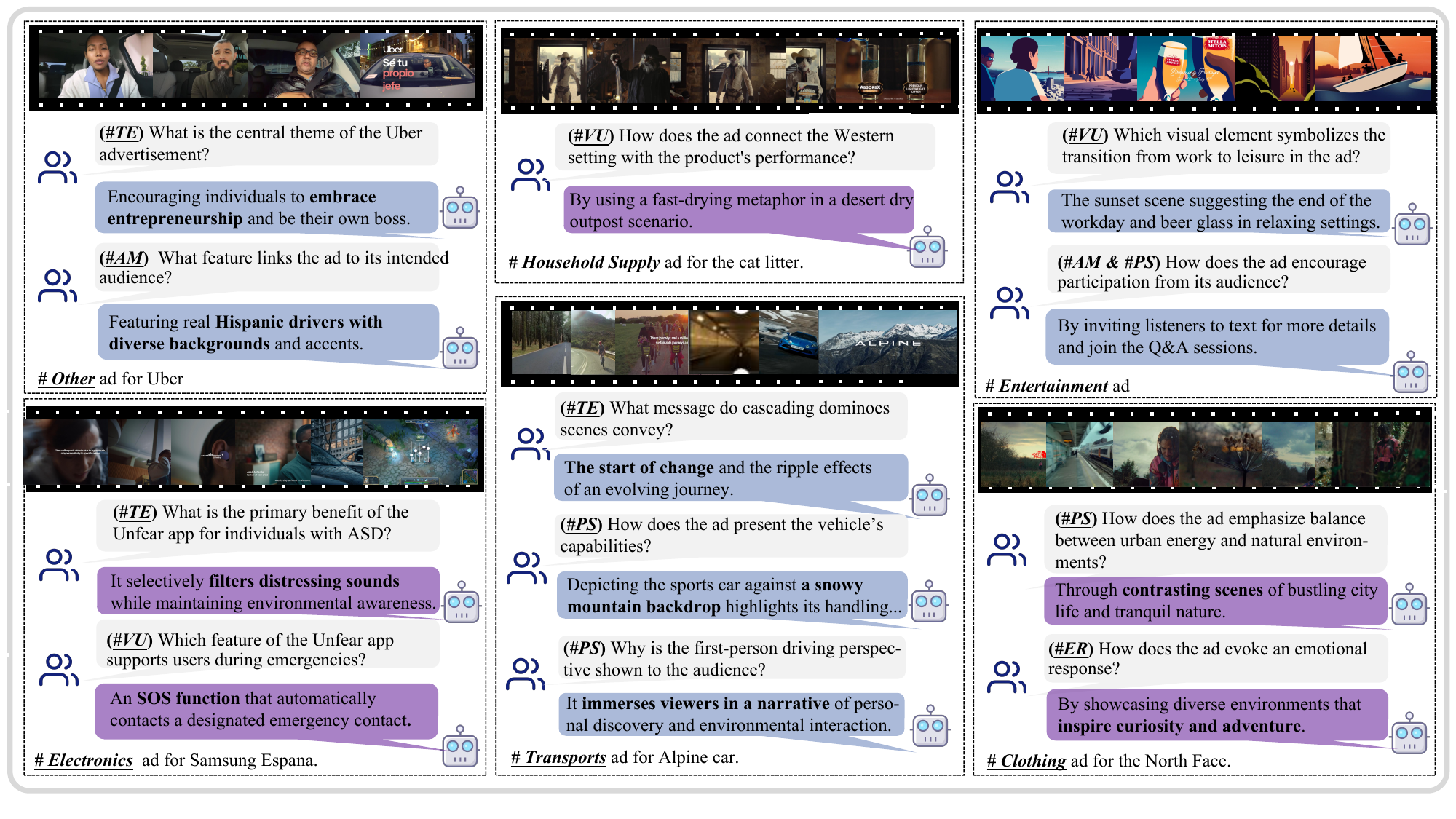} 
\caption{Examples of \texttt{AdsQA}.}
\label{fig_2}
\end{figure*}

We design a role-played multi-agent annotation framework for our three sequential stages, as illustrated in Fig.~\ref{fig_1}.

\textbf{In the initial stage}, a master agent is recruited to generate preliminary QA annotations for the given modality-interleaved sequence $V$. The master agent analyzes the input sequence and produces an initial set of question-answer pairs covering the key elements of the video content.

The framework then enters an \textbf{iterative refinement stage} where the master agent systematically evaluates the quality of existing annotations. This evaluation determines whether to recruit specialized expert agents to improve specific aspects of the annotations. When expert agents are needed, the master agent selects appropriate profiles matching the required expertise and instantiates them accordingly. Each recruited expert agent generates new QA annotations leveraging its specialized knowledge, which the master agent then incorporates into the existing set through careful revision and integration. This iterative process continues until the master agent determines the annotations have reached sufficient quality.

\textbf{In the final stage}, the master agent synthesizes all annotations produced during the iterations into a cohesive final set of QA pairs. This synthesis process combines the breadth of coverage from the initial annotations with the depth of specialized knowledge from expert contributions, resulting in comprehensive and accurate video understanding annotations ready for benchmark use.

We use both commercial and open-source large language models, including GPT-4, GPT-4o, LLaVA-OV-72B, and Qwen2.5-VL-72B. Based on comprehensive evaluations of QA pair annotation quality, cost efficiency, and data style consistency, we ultimately selected Qwen2.5-VL-72B for our final dataset generation. The automated prompts used in this process are detailed in Tab.~\ref{tab:instruction_format_1}.

\subsection{Postprocessing}

\noindent \textbf{Video Domain Classification.}
We leverage the hashtags collected along with the videos to classify the ad domains. 
After clustering all domain-related hashtags, we observe that the majority of ads fall into eight primary categories, including Food, Clothing, Health, Household Supply, Public Service, Entertainment, Transport, and Electronics. 
Advertisements that fail to be clearly categorized (e.g., those labeled as ``public interest'' without a specific domain assignment) are grouped into an ``Others''. The full list of our domain taxonomy is in Tab.~\ref{tab:domains}.

We note that some advertisements involves multiple domains (\textit{e.g.}, cross-brand collaborations between different industries). However, for consistency, we only retain the most relevant primary domain for each advertisement.

\noindent \textbf{Question Type Classification.} We do not predefine question types during generation but instead perform post-hoc classification, as not all videos are suitable for generating every type of question. 
Some questions are permitted to be classified into two categories, because they may involve multiple aspects of advertisements (\textit{e.g.}, how visual elements interact with the theme).
Each QA pair is classified three times using GPT-4o, with manual verification to resolve inconsistencies or non-type cases, ensuring annotation quality. The prompts are listed in Tab.~\ref{tab:instruction_format_5}.

\noindent \textbf{Human Check and Annotator Team.}
Five human annotators from our team designed the agent profiles and prompts, conducted qualitative analysis, and performed human evaluation. 
In the first stage, each annotator is required to view the advertisement video and select 3-7 QA pairs from the candidate pool within approximately 5 minutes. If modifications are deemed necessary, the annotator will review both the questions and the answers.
In the second stage, we recruited two additional annotators to review the modifications and correct any remaining errors.

\noindent \textbf{Evaluation Metrics.} We follow their framework and employ GPT-4o to assist in evaluating free-form text similarity. Our prompts used for model-based evaluation are listed in Tab.~\ref{tab:instruction_format_4}.

\section{Benchmark Analysis}
Fig. \ref{fig_2} shows six video examples randomly selected from the domain of ``others'', involving ``household supply'', ``entertainment'', ``electronics'',  ``transports'', and ``clothing''. 
\textcolor{black}{We observe that:
1) The videos used in the AdsQA dataset are aesthetically appealing and of high quality;
2) AdsQA focuses on intention-driven video understanding, but traditional benchmarks emphasize perceptual-level comprehension; For example, a question from the well-known ActivityQA dataset~\cite{DBLP:conf/aaai/YuXYYZZT19}, ``What color are the gloves worn by the person who is skiing?'' is answered with the word ``black''.
3) AdsQA is a comprehensive benchmark capable of evaluating Video-LLMs across multiple dimensions, including visual cues, emotions, themes, persuasive strategies, and user modeling in advertisements.}

\begin{figure}[t]
\centering
\includegraphics[width=0.5\textwidth]{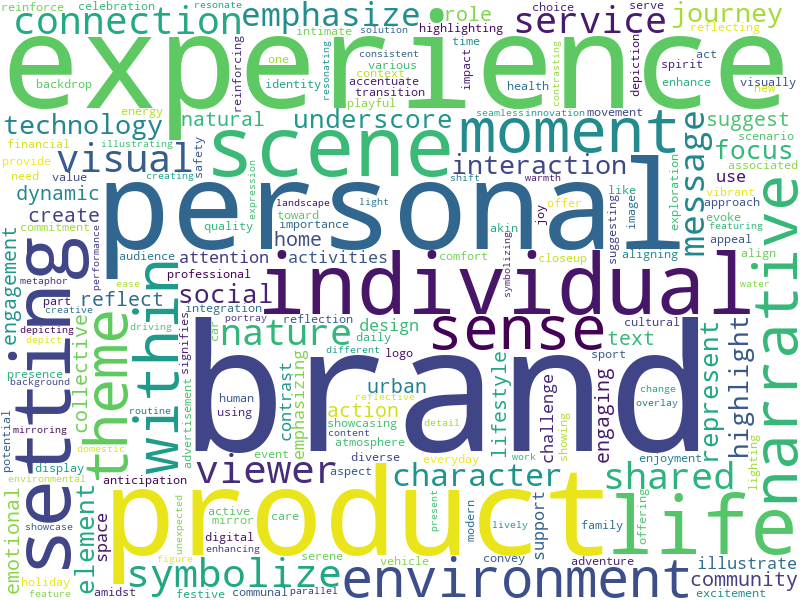} 
\caption{Word cloud of the questions of \texttt{AdsQA}.}
\label{fig_wc}
\end{figure}

\begin{figure}[t]
\centering
\includegraphics[width=0.5\textwidth]{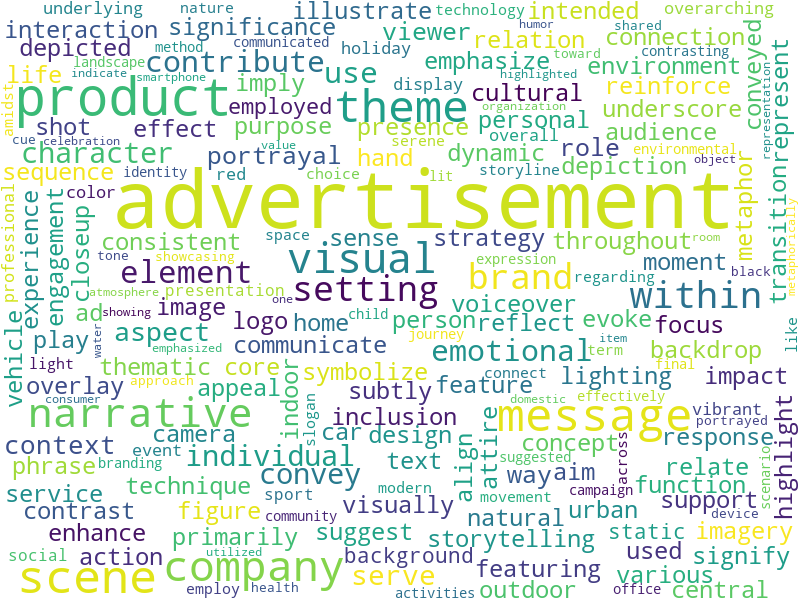} 
\caption{Word cloud of the ground-truth answers of \texttt{AdsQA}.}
\label{fig_wc_ans}
\end{figure}

Fig.~\ref{fig_wc} and Fig.~\ref{fig_wc_ans} are two word clouds of \texttt{AdsQA} questions and answers, respectively.
Tab.~\ref{tab_bench} presents a comparative overview of the AdsQA Benchmark with both general-purpose and domain-specific video understanding datasets.
Compared to these datasets, the AdsQA Benchmark provides a comparable scale in terms of video duration, number of videos, and number of QA pairs.
In a nutshell, the AdsQA Benchmark offers unique advantages, including the public availability of videos, diverse and comprehensive QA pairs, and high data quality.
\begin{table*}[]
\centering
\small
\begin{tabular}{ccccccc}
\toprule
{\bf Datasets} {(test)} & {\bf Domains}     & {\bf Annotation}  & {\bf Avg. Length} & {\bf QA Pairs} & {\bf Task Types} & {\bf Access} \\ \midrule
MSRVTT-QA~\cite{xu2017video}      & Open       & Auto        & 15s          & 72,820   & Open         & $\checkmark$          \\
Pitt~\cite{DBLP:conf/cvpr/HussainZZYTAOK17}           & Ads        & Manual+Auto & unknown      & unknown & Open      &            \\
TVQA~\cite{DBLP:conf/emnlp/LeiYBB18}           & Movie      & Manual      & 160s         & 15,253   & MCQs         &            \\
How2QA~\cite{sanabria2018how2}         & Open       & Manual      & 60s          & 4,400    & Open         & $\checkmark$          \\
ActivityNet-QA~\cite{DBLP:conf/aaai/YuXYYZZT19} & Open       & Manual      & 180s         & 8,000    & Open         & $\checkmark$          \\
VideoBench~\cite{DBLP:journals/corr/abs-2311-16103}        & Open       & Manual      & 56s          & 4,000    & MCQs       & $\checkmark$          \\
EgoSchema~\cite{DBLP:conf/nips/MangalamAM23}      & Egocentric & Auto        & 180s         & 5,031    & MCQs       & $\checkmark$          \\
MVBench~\cite{DBLP:conf/cvpr/0002WH00LWX0L0024}        & Open       & Manual      & 15s          & 4,000    & MCQs       & $\checkmark$          \\
\midrule
\texttt{AdsQA}    & Ads        & Manual+Auto & 52.9s          & 7,859     & Open       & $\checkmark$           \\ \bottomrule
\end{tabular}
\caption{Comparison between our \texttt{AdsQA} and other Video QA benchmarks. {``MCQs'' denotes ``multiple-choice questions''.}}
\label{tab_bench}
\end{table*}

\section{Implementation Details}
\label{sec_dem}
\subsection{Details of Baselines}
\noindent \textbf{Human Performance.} We employed five human evaluators to assess the same randomly sampled subset (40 videos, 200 QA pairs). Each evaluator was instructed to provide answers based solely on the given questions and video content, without access to any additional background information (\textit{e.g.}, metadata).

\noindent \textbf{GPT-4o}~\cite{achiam2023gpt} \textcolor{black}{is one of the most famous commercial MLLMs, capable of processing both text and image inputs within a unified framework.} In our work, we first utilize Qwen2.5-VL-72B~\cite{bai2025qwen2} to generate captions for advertising videos, then feed these captions as contextual information (instead of raw visual content) into GPT-4o for question answering.

\noindent \textbf{Gemini-2.5-Pro}~\cite{comanici2025gemini}  \textcolor{black}{is the current state-of-the-art commercial MLLM, excelling in unified text/image/audio/video understanding, million-token context processing, and human-like reasoning for coding and complex tasks.} Due to budget constraints, we randomly selected a subset of 600 samples from the dataset. We directly encoded both the audio and video streams into the Gemini-2.5-Pro model for question answering.

\noindent \textbf{VideoLLaMA-2}~\cite{DBLP:journals/corr/abs-2406-07476} introduces VideoLLaMA 2, a set of Video-LLMs designed to enhance spatial-temporal modeling and audio understanding in video and audio tasks.

\noindent \textbf{Intern\_XComp}~\cite{internlmxcomposer2_5} presents InternLM-XComposer-2.5 (IXC-2.5), a versatile MLLM that supports long-contextual input and output, achieves GPT-4V level capabilities with a 7B LLM backbone.

\noindent \textbf{MiniCPM-o 2.6}~\cite{yao2024minicpm} is the latest model in the MiniCPM-V series of edge-side MLLMs. The models can take images, video, text, and audio as inputs and provide high-quality text and speech outputs in an end-to-end fashion.

\noindent \textbf{Qwen2-VL}~\cite{DBLP:journals/corr/abs-2409-12191} based on Qwen2, is an advanced MLLM that can understand images of different resolutions and aspect ratios, achieving leading performance on existing video benchmarks.

\noindent \textbf{LLaVA-Video}~\cite{zhang2024videoinstructiontuningsynthetic} introduces LLaVA-Video-178K, a high-quality synthetic dataset for video instruction-following tasks, and presents LLaVA-Video, a new video LMM trained on this dataset and existing visual instruction data, demonstrating strong performance across various video benchmarks.

\noindent \textbf{LLaVA-Onevision}~\cite{li2024llava} is an open-sourced MLLM that achieves competitive performance in various computer vision tasks, including single-image, multi-image, and video scenarios.
It further demonstrates robust transfer learning capabilities across modalities and scenarios.

\noindent \textbf{Qwen2.5-VL-7B/72B}~\cite{bai2025qwen2} is currently the state-of-the-art open-sourced MLLM, built upon the Qwen2.5 backbone. It supports text, images, and videos as input. Moreover, Qwen2.5-VL-72B outperforms commercial MLLMs such as GPT-4o on multiple multimodal benchmarks.




\subsection{Prompt Templates}
Tab~\ref{tab:instruction_format_1}  and Tab~\ref{tab:instruction_format_4} present the prompt templates utilized in our model, wherein $<$Video$>$ and $\{*\}$ denote placeholders for video and text content, respectively. 

\subsection{Hyperparameters}
In this study, we utilized four A100 GPUs for training the GRPO model and optimized it using a series of hyperparameters.  During training, the batch size per device was set to 1, with a gradient accumulation step of 1 to optimize computational resource utilization. The learning rate was set to 1e-6, and we adopted BF16 precision to improve computational efficiency and reduce memory consumption. FlashAttention-2 was employed to accelerate training and enhance model performance. Additionally, the maximum number of pixels was set to 401,408 to accommodate high-resolution input data. To control gradient explosion and stabilize training, the maximum gradient norm was set to 20, and the model was trained for two epochs to ensure adequate learning of the data distribution. Furthermore, in the generation task, we set the number of generated outputs to 8 to enhance the model's generalization ability and output diversity.

\section{Term of Use}
Our benchmark collects videos from a publicly accessible, creator-uploaded ad video website, strictly adhering to the requirements of its terms of use. The website is a non-commercial platform allowing free access and downloads for non-commercial purposes. Moreover, the website's mission to inspire creativity and foster the exchange of ideas aligns closely with our objective of enable AI understand creative content.
\textbf{Given copyright considerations, we will release download scripts, along with the URLs and features of the videos, similar to other video benchmarks}~\cite{DBLP:conf/aaai/YuXYYZZT19,DBLP:journals/corr/abs-2406-10221,DBLP:journals/corr/abs-2410-10818}.
Before release, all Q\&A annotations have been reviewed to ensure they do not contain any information that identifies individuals by name or any offensive content.

\section{Limitations}
Our work introduces the first LLM benchmark for advertising video understanding and proposes ReAd-R, a DeepSeek-R1 styled model. However, several limitations remain:
1) We carefully removed videos appearing in existing Video-LLM training corpora from AdsQA, but we cannot guarantee complete exclusion from all LLM pretraining data.
2) Potential human bias may exist due to variations among annotators during data collection.
3) The majority of QA pairs were automatically generated by LLMs, despite manual revision of some samples. This may introduce potential biases inherent to large language models.
4) Despite careful prompt design, model-based auto-evaluation faces common challenges where results may not always align with human assessment.
We will address these issues in future work.

\section{Workshop and Challenge}
We organize the ICCV 2025 MARS2 Workshop ``Multimodal Reasoning and Slow Thinking in Large Model Era: Towards System 2 and Beyond''. Our \texttt{AdsQA} is used to support a competition track as its testset. 

\begin{table*}[htbp]
\small
\centering
\begin{tabular}{p{15cm}l}
\toprule
\rowcolor{gray!20} \textbf{\textit{Initial Q\&A Pairs Generation.}} \\
\midrule
You are tasked with generating questions from advertisement descriptions. Use the description alone to craft the questions and avoid making assumptions. The correct answer should blend seamlessly with the wrong ones in terms of length and complexity. Do not use direct quotes, and keep terminology simple. Questions should relate only to the content of the advertisement and avoid any external or behind-the-scenes details.This advertisement contains the following \{\#Scene\_nums\} scenes:

\{\#Scene\_Descriptions\}

Voiceover: \{\#Voiceover\}

The above scene descriptions are from the \{\#Brand\} advertisement, titled "\{\#Title\}". Some potentially useful information about the Theme and the Brand is as follows:\\
Theme, Brand and Product Features: \{\#Theme\}.\\
Create No more than ten  questions based on this synopsis.
Use these aspects as a reference when asking questions:\\
1. Theme and core message (compulsory)\\
2. Conveyance method for Theme, Brand, and Product features (if applicable)\\
3. Specific visual elements (object, person, scene, event, etc.) and their relation to the theme (No more than three questions)\\
4. Specific detail's connection to the overall theme (compulsory)\\
5. Target audience characteristics (if applicable)\\
6. Emotional impact and tactics used\\
7. Storyline and narration (if applicable)\\
8. Metaphors or humorous techniques (if present)\\
9. Logical arguments, factual claims, or expert opinions (if present)\\
10. Characters and their relevance to the theme and audience (if present)\\
11. Creativity and the overall impression of the ad. (if applicable)\\
For each question, provide only one correct answer. The answers must be unique, and unbiased.  Print each correct answer exactly as 'Correct answer: [full answer]'. \\
\midrule
\midrule
\rowcolor{gray!20} \textbf{\textit{Call for Expert Agents}} \\
\midrule
\{\#Initial\_Prompt\} \\
\{\#Initial\_Annotation\} \\
Now, you can create and collaborate with multiple experts to improve your generated question-answer pairs. Therefore, please describe in as much detail as possible the different skills and focuses you need from multiple experts individually.
We will provide each expert with the same information and query. However, please note that each profession has its own specialization, so you can assign each expert to just one sub-task to ensure a more refined response.
We will relay their responses to you in turn, allowing you to reorganize them into a better generation.
Please note that the description should be narrated in the second person, for example: You are a XXX.

These are the descriptions of the experts you have created before for this task:

Therefore, please remember you should not repeatedly create the same experts as described above.
Now, you can give the description for a new expert (Please note that only be one, do not give multiple at one time): \\
\midrule
\midrule
\rowcolor{gray!20} \textbf{\textit{Profiles for Expert Agents}} \\
\midrule
 \{\# You are a conservation psychologist, specializing in understanding and promoting the psychological and emotional connection people have with nature and wildlife. Your expertise includes analyzing how visual and textual messaging in media can influence individuals' attitudes and behaviors towards conservation and environmental protection. Your focus lies in interpreting the emotional responses elicited by multimedia content and identifying the aspects of an advertisement that enhance the viewers' sense of urgency or empathy towards the subject. You provide insights on the psychological impact of specific scenes, colors, narratives, and the use of statistics or facts in fostering a sense of environmental stewardship and activism. Your role is to evaluate the effectiveness of the environmental messages conveyed and suggest ways to strengthen the emotional appeal and call to action within the advertisement.\} \\
\midrule
\midrule
\rowcolor{gray!20} \textbf{\textit{The Role-played Agent Prompt for Annotation Generations}} \\
\midrule
\{\# Profile\}\\
\{\# Initial Prompt\}\\
\midrule
\midrule
\rowcolor{gray!20} \textbf{\textit{Master Agent Prompt for Annotation Revision}} \\
\midrule
\{\# Current QA Annotations or Initial Annotations\}\\
You invite an expert whose description is: \{\# Profile\}\\
\{\# QA Annotations Generated by the last Expert Agent\}\\
Now you can refine your question-answer pairs with his generation to create more professional and challenging question-answer pairs.
Keep in mind that his generation may not be perfect, so critically decide whether to accept some parts of his response or stick with your original one.
Revised Question-Answer Pairs:\\
\bottomrule
\end{tabular}
\small
\caption{Prompt Format for the Annotation Generation.}
\label{tab:instruction_format_1}
\end{table*}

\begin{table*}[htbp]
\small
\centering
\begin{tabular}{p{15cm}l}
\toprule
\rowcolor{gray!20} \textbf{\textit{Prompts for Model-based Evaluation (Strict Acc).}} \\
\midrule
 You are an advertising expert specializing in evaluating whether a respondent's answer after watching a video matches the golden answer. We will provide the video's Meta-Information, Question, Golden Answer, and the Response to be judged below.\\
\#\#\#The meta-information includes the advertisement video's theme, creative points, and a brief content description, which can be regarded as ground-truth information, as follows::
\{\#meta\_info\}

\#\#\#Question: 
\{\#question\}

\#\#\#Golden Answer: 
\{\#golden\_answer\}

\#\#\#Rule:\\
1. If the response to be judged contains ALL key information of the golden answer or expresses the same meaning using other sentences or synonyms, it is considered a match with the golden answer, and the output is 1.\\
2. If the response to be judged does NOT contain the key information from the golden answer, it is considered a mismatch, and the output is 0.\\
3. The response to be judged should NOT contain any content that is contradictory, conflicting, or unreasonable when inferred from the meta-information. If such content exist, it is considered a mismatch, and the output is 0.\\

\#\#\#Response to be judged: \\
{response}

\#\#\#Instructions:\\
Follow the format below and do not give any extra outputs:\\
Answer: 0 (if the response does not match)\\
Answer: 1 (if the response matches) \\
\midrule
\midrule
\rowcolor{gray!20} \textbf{\textit{Prompts for Model-based Evaluation (Relexed Acc).}} \\
\midrule
You are an advertising expert specializing in evaluating whether a respondent's answer after watching a video matches the golden answer. We will provide the video's Meta-Information, Question, Golden Answer, and the Response to be judged below.\\
\#\#\#The meta-information includes the advertisement video's theme, creative points, and a brief content description, which can be regarded as ground-truth information, as follows::
\{\#meta\_info\}

\#\#\#Question: 
\{\#question\}

\#\#\#Golden Answer: 
\{\#golden\_answer\}

\#\#\#Rule:\\
1. If the response to be judged contains ALL key information of the golden answer or expresses the same meaning using other sentences or synonyms, it is considered a match with the golden answer, and the output is 1.\\
2. If the response to be judged does NOT contain the key information from the golden answer, it is considered a mismatch, and the output is 0.\\
3. The response to be judged should NOT contain any content that is contradictory, conflicting, or unreasonable when inferred from the meta-information. If such content exist, it is considered a mismatch, and the output is 0.\\
4. If the response to be judged contains the MOST of key information of the golden answer and, do NOT contain any information that is contradictory, conflicting, or unreasonable when inferred from the meta-information, it is considered a partial match, and the output is 0.5.

\#\#\#Response to be judged: \\
{response}

\#\#\#Instructions:\\
Follow the format below and do not give any extra outputs:\\
Answer: 0 (if the response does not match)\\
Answer: 0.5 (if the response partially match)\\
Answer: 1 (if the response matches)\\


\bottomrule
\end{tabular}
\small
\caption{Prompt Templates for Model-based Evaluation}
\label{tab:instruction_format_4}
\end{table*}

\begin{table*}[htbp]
\small
\centering
\begin{tabular}{p{15cm}l}
\toprule
\rowcolor{gray!20} \textbf{\textit{Prompts for Question Classification.}} \\
\midrule

Classify the question-answer pair into one of the following categories :\\
Type\_1: The question-answer pair that focuses on the visual concepts, such as video details, characters in videos, a certain object, a certain scene, slogans presented in video, events, plot, and their interaction.\\
Type\_2: The question-answer pair that emotional content by ad videos and assesses the potential psychological impact of these emotions.\\
Type\_3: The question-answer pair that focuses on the brand value, goal, theme, underlying message, or central idea that the ad explores and conveys.\\
Type\_4: The question-answer pair that focuses on persuasion strategies that ad videos convey their core messages. These messages may not be directly articulated but could instead engage viewers through humor and visual rhetoric. (\textit{e.g.}, Any questions about the symbols, metaphors, humor, exaggeration, and any questions that focus on the Logical arguments, factual claims, Statistical charts, or expert opinions. (Any question about presenting factual information and logical arguments to demonstrate the product’s benefits and value)\\
Type\_5: The question-answer pairs focus on the engagement, call for action, target audience, the characteristics of the target demographic, and who will be engaged.\\ 
If this question could belong to multiple categories, please choose the most relevant two.\\
Question: \{\#question\}\\
Answer: \{\#answer\}\\
Your output should be just one or two of Type\_1, Type\_2, Type\_3, Type\_4, Type\_5, and nothing else. \\
\bottomrule
\end{tabular}
\small
\caption{Prompt Templates for Question Classification}
\label{tab:instruction_format_5}
\end{table*}

\begin{table*}[htbp]
\small
\centering
\begin{tabular}{p{15cm}l}
\toprule
\rowcolor{gray!20} \textbf{\textit{The Prompt Template for Qwen2-VL, Qwen2.5-VL, MiniCPM-o 2.6, InternLM-XComposer2d5, and VideoLLaMA2.}} \\
\midrule
Video: \{\#Frames\}

Voiceover: \{\#Voiceover\}
By watching the video, you are required to answer a question within 30 words: \\
Question: \{\#question\}\\
\midrule
\midrule
\rowcolor{gray!20} \textbf{\textit{The Prompt Template for LLaVA-Onevision and LLaVA-Video.}} \\
Video: \{\#Frames\} \\
The advertisement video lasts for \{\#video\_time\:\.2f\} seconds, and \{\#frame\_nums\} frames are uniformly sampled from it. These frames are located at \{\#frame\_time\}.\\
Voiceover: \{\#Voiceover\} \\
By watching the video, you are required to answer a question within 30 words: \\
Question: \{\#question\}\\
\midrule
\rowcolor{gray!20} \textbf{\textit{The Prompt Template for Chain-of-Thought Baselines. (Gemini-2.5-pro)}} \\
\midrule
Voiceover: \{\#Voiceover\} \\
Question: \{\#question\} \\
Analyze the provided video and voiceover (if present) to perform step-by-step reasoning. Structure your output as follows:\\
1. Chain-of-Thought (CoT): Enclose detailed reasoning in \textless think\textgreater tags, covering key visual/audio cues, logical connections, and deductions. \\
2. Final Answer: Provide a concise answer which answers the given question (less than 30 words) in \textless answer\textgreater tags. \\

Example Output: \\
\verb|```| \\
\textless think\textgreater
The video shows a crowded street with festival decorations (red lanterns, fireworks)......
\textless /think\textgreater \\
\textless answer\textgreater A Chinese New Year festival is being celebrated.\textless /answer\textgreater \\
\verb|```| \\
\bottomrule
\end{tabular}
\small
\caption{Prompt Format for Inference.}
\label{tab:instruction_format_6}
\end{table*}

\end{document}